%% file: MAIN.tex
\documentclass[runningheads]{llncs}
\usepackage[T1]{fontenc}
\usepackage{graphicx,verbatim}
\usepackage{booktabs}
\usepackage{hyperref}
\usepackage{subcaption}
\usepackage{xspace}
\usepackage[table]{xcolor}
\usepackage{amsmath}

\DeclareMathOperator*{\argmin}{arg\,min}
\newcommand{\VmriPLA}[0]{98.30}
\newcommand{\VmriC}[0]{99.58 $\pm$ 3.19}
\newcommand{\VmriTEA}[0]{87.80}
\newcommand{\VmriLEA}[0]{94.48}
\newcommand{\VctPLA}[0]{99.18}
\newcommand{\VctC}[0]{99.63 $\pm$ 4.37}
\newcommand{\VctTEA}[0]{96.30}
\newcommand{\VctLEA}[0]{97.22}
\newcommand{\PLP}[0]{PL perc.\xspace}
\newcommand{\subjcorr}[0]{sub. corr.\xspace}
\newcommand{\TEAacc}[0]{TEA rec.\xspace}
\newcommand{\LEAacc}[0]{LEA rec.\xspace}

\newcommand{\gcmidrule}[1]{\arrayrulecolor{black!30}\cmidrule{2-#1}\arrayrulecolor{black}}
\newcommand{\gccmidrule}[2]{\arrayrulecolor{black!30}\cmidrule{#1-#2}\arrayrulecolor{black}}
\begin{document}
\title{VERIDAH: Solving Enumeration Anomaly Aware Vertebra Labeling across Imaging Sequences}
%

\author{PUBLISHED IN EUROPEAN RADIOLOGY \url{https://link.springer.com/article/10.1007/s00330-026-12966-x}\\-\\Hendrik Möller$^{1,2}$, Hanna Schoen$^{1, 5}$, Robert Graf$^{1,2}$, Matan Atad$^{1,2}$, Nathan Molinier$^{3}$, Anjany Sekuboyina$^{4}$, Bettina K. Budai$^{7}$, Fabian Bamberg$^{8}$, Steffen Ringhof$^{8}$, Christopher Schlett$^{8}$, Tobias Pischon$^{9}$, Thoralf Niendorf$^{10}$, Josua A. Decker$^{11}$, Marc-André Weber$^{5}$, Bjoern Menze$^{4}$, Daniel Rueckert$^{2, 6}$, and Jan S. Kirschke$^{1}$}  
\authorrunning{Möller et al.}
\institute{%
$[1]$ Department for Interventional and Diagnostic Neuroradiology, TUM University Hospital \\
$[2]$ Chair for AI in Healthcare and Medicine, Technical University of Munich (TUM)\\
$[3]$ NeuroPoly Lab, Institute of Biomedical Engineering, Polytechnique Montreal, and Mila -- Quebec AI Institute\\
$[4]$ Department of Quantitative Biomedicine, University of Zurich\\
$[5]$ Department of Diagnostic and Interventional Radiology, Pediatric Radiology and Neuroradiology, Rostock University Medical Center\\
$[6]$ Department of Computing, Imperial College London\\
$[7]$ Department of Diagnostic and Interventional Radiology, University Hospital Heidelberg\\
$[8]$ Department of Diagnostic and Interventional Radiology, Medical Center -- University of Freiburg\\
$[9]$ Molecular Epidemiology Research Group, Max-Delbrueck-Center for Molecular Medicine in the Helmholtz Association (MDC)\\
$[10]$ Berlin Ultrahigh Field Facility (B.U.F.F.), Max Delbrueck Center for Molecular Medicine\\
$[11]$ Department of Diagnostic and Interventional Radiology, University Hospital Augsburg
}%
  
\maketitle              
\begin{abstract}
The human spine commonly consists of seven cervical, twelve thoracic, and five lumbar vertebrae. However, enumeration anomalies may result in individuals having eleven or thirteen thoracic vertebrae and four or six lumbar vertebrae. Although the identification of enumeration anomalies has potential clinical implications for chronic back pain and operation planning, the thoracolumbar junction is often poorly assessed and rarely described in clinical reports. Additionally, even though multiple deep-learning-based vertebra labeling algorithms exist, there is a lack of methods to automatically label enumeration anomalies. Our work closes that gap by introducing "Vertebra Identification with Anomaly Handling" (VERIDAH), a novel vertebra labeling algorithm based on multiple classification heads combined with a weighted vertebra sequence prediction algorithm. We show that our approach surpasses existing models on T2w TSE sagittal (\VmriPLA\% vs. 94.24\% of subjects with all vertebrae correctly labeled, $p<0.001$) and CT imaging (\VctPLA\% vs. 77.26\%  of subjects with all vertebrae correctly labeled, $p<0.001$) and works in arbitrary field-of-view images. VERIDAH correctly labeled the presence of thoracic enumeration anomalies in \VmriTEA \% and \VctTEA \% of T2w and CT images, respectively, and lumbar enumeration anomalies in \VmriLEA \% and \VctLEA \% for T2w and CT, respectively. 
Our code and models are available at: \href{https://github.com/Hendrik-code/spineps}{https://github.com/Hendrik-code/spineps}.%


\end{abstract}

\input{LaTeX/m_introduction}
\input{LaTeX/m_methodology}
\input{LaTeX/m_results}
\input{LaTeX/m_ablations}
\input{LaTeX/m_discussion}
\input{LaTeX/m_appendix}
\bibliographystyle{splncs04}
\bibliography{LaTeX/bib}
\end{document}

%% file: LaTeX/m_introduction.tex
\section{Introduction}
\label{sec:introduction}

The benefits of automated processing and diagnosis of medical images have often been demonstrated \cite{sim2020deep,kickingereder2019automated,de2018clinically}. For the analysis of the spine, these automated assessment systems usually require accurate identification of the vertebrae, such as for fracture prediction \cite{loffler2020vertebral}. The human spine typically consists of seven cervical, twelve rib-bearing thoracic, and five lumbar vertebrae \cite{thawait2012spine}. 
However, there can be enumeration anomalies, causing a deviation of the number of thoracic and/or lumbar vertebrae by one \cite{tatara2021changes,wigh1980thoracolumbar}. These thoracic enumeration anomalies (TEA) and lumbar enumeration anomalies (LEA) are often accompanied by transitional anomalies, such as thoracolumbar transitional vertebrae (TLTV). These TLTV are vertebrae with hybrid morphological features from thoracic and lumbar vertebrae \cite{park2016thoracolumbar}.
Studies have shown that the presence of enumeration anomalies is strongly correlated with transitional ones and has potential clinical implications for chronic back pain and degenerative changes \cite{nardo2012lumbosacral,apaydin2019lumbosacral}. Ignoring enumeration anomalies can also lead to vertebrae being labeled incorrectly. This can result in wrong-site surgery, where the intended vertebra is missed and adjacent vertebrae are operated on by mistake \cite{tatara2021changes}.

A naive approach to identifying vertebrae is to count them in a cranial-to-caudal or caudal-to-cranial order. This requires a known reference point within the field of view (FOV), such as the first thoracic vertebra (identified by the presence of a rib) or the sacrum (to identify the last lumbar vertebra). If either is visible, the vertebrae can be counted and labeled accordingly. However, CT and MRI scans often have a limited FOV along the superior-inferior direction, making manual assessments inconsistent. Furthermore, anatomical variations, such as vertebral enumeration anomalies, render the counting approach error-prone and make both counting directions incompatible. For example, even if the last lumbar vertebra is visible, simple counting cannot distinguish between spines with four, five, or six lumbar vertebrae.

The automatic identification of vertebrae has been the subject of dedicated research \cite{liao2018joint,payer2019integrating,huang20213d,sekuboyina2021verse,tao2022spine,raja2010labeling,cai2015multi}.
There have been many development efforts, especially in CT \cite{liao2018joint,payer2019integrating,huang20213d}. However, some approaches are only tested on small cohorts \cite{kawathekar2024novel}, and almost none of them consider TEA or LEA, as the existing systems can only predict a maximum of 12 thoracic and five lumbar vertebrae \cite{windsor2022spinenetv2}. We found three automated approaches that theoretically take this into account. Sekuboyina et al. \cite{sekuboyina2023pushing} developed a method for localization and identification of vertebrae in CT, using a conditional random field and user input to refine errors. Meng et al. \cite{meng2022vertebrae} utilized segmentation masks of vertebrae and a graph optimization problem for vertebra identification. Finally, Warszawer et al. \cite{yehudawarszawer202514920281} locate specific intervertebral discs during the task of segmentation in order to identify the sequence of vertebrae visible in the image by counting from those discs in both directions. However, these methods either fail to label TEA and LEA cases reliably and robustly in arbitrary FOVs or have never been validated on a large dataset.

Seizing this opportunity, we use large cohorts of 3D MRI and CT images and establish a robust labeling method that can be used even in the presence of enumeration anomalies and works in arbitrary FOVs. We propose to use a classifier with multiple classification heads to receive vertebra-label related outputs. Then, we use those outputs jointly to solve a constrained optimization problem in order to find the sequence of vertebra labels where the agreement of the multiple vertebrae and classification outputs is highest.

The contributions of this study can be summarized as follows: (1) establish a new multi-classification head strategy of training vertebra labeling classifiers, (2) introduce a label sequence predictor based on the classifier outputs, (3) show the superiority of our method to the current state-of-the-art on a large cohort of data on the CT and T2w MRI sequence, (4) make our approach and model weights publicly available.

%% file: LaTeX/m_methodology.tex
\section{Material and Methods}
\label{sec:method}

\subsection{Data}
We utilize two data cohorts, the external German National Cohort (NAKO) \cite{bamberg2015whole,DAE240963} and an in-house CT dataset. Informed consent was waived for this retrospective study (593/21 S-NP).

\begin{table}[htbp]
    \centering
    \caption{Demographics for the whole datasets and individual train and test splits. Abbreviations: yrs: years, kg: kilogram, m: meters, M.: Mean, SD: standard deviation, TEA: Thoracolumbar Enumeration Anomaly, LEA: Lumbar Enumeration Anomaly.}
    \label{tab:nakodemo}
    \setlength{\tabcolsep}{3pt}
    \begin{tabular}{rccc}
        \toprule
        \textbf{NAKO T2w sagittal}        & Summary   & Train         & Test \\
        \midrule
        Number of Scans             & 6291          & 5291          & 1000 \\
        Age Range (yrs)         & 20 -- 72      & 20 -- 72       & 22 -- 72 \\
        M. Age (yrs) $\pm$ SD   & $52 \pm 11$   & $52 \pm 12$   & $52 \pm 11$ \\
        M. Weight (kg) $\pm$ SD & $80 \pm 16$   & $80 \pm 16$   & $80 \pm 16$ \\
        M. Height (m) $\pm$ SD    & $1.7 \pm 0.1$ & $1.7 \pm 0.1$ & $1.7 \pm 0.1$ \\
        Sex ($\%$ female)       & 52.4          & 52.4          & 52.1 \\
        TEA Rate ($\%$)         & 5.8          & 5.4          & 8.2 \\
        LEA Rate ($\%$)         & 9.7          & 8.8          & 14.6 \\
        M. Visible Vertebrae $\pm$ SD & $23.1 \pm 0.3$ & $23.1 \pm 0.3$ & $23.1 \pm 0.4$ \\
        \midrule
        \textbf{CT Cohort}                   & Summary       & Train         & Test \\
                \midrule                
        Number of Scans             & 1536          & 1171          & 365 \\
        Age Range (yrs)         & 18 -- 100      & 18 -- 100       & 20 -- 100 \\
        M. Age (yrs) $\pm$ SD   & $74 \pm 21$   & $74 \pm 21$   & $74 \pm 21$ \\
        M. Weight (kg) $\pm$ SD & $76 \pm 16$   & $75 \pm 14$    & $81 \pm 21$ \\
        M. Height (m) $\pm$ SD    & N/A & N/A & N/A \\
        Sex ($\%$ female)       & 49.1          & 50.9          & 44.3 \\
        TEA Rate ($\%$)         & 5.5          & 5.1          & 6.8 \\
        LEA Rate ($\%$)         & 7.3          & 7.1          & 7.9 \\
        M. Visible Vertebrae $\pm$ SD & $16.7 \pm 3.6$ & $16.7 \pm 3.5$ & $16.8 \pm 4.0$ \\
        \bottomrule
    \end{tabular}
\end{table}%

For the T2w sagittal MRI data, we utilized a random subset of 6291 subjects of the German National Cohort (NAKO) and split the data roughly into 70/15/15\% train/validation/test (see Table \ref{tab:nakodemo}). We utilized SPINEPS \cite{moller2024spineps}, a segmentation network, to derive instance segmentation annotations for all vertebrae in our T2w sagittal MRI data. We excluded 33 subjects (0.52\%) where this segmentation failed. We did not observe obvious reasons for these failures, such as hyperintense spots or pathologies. The segmentations yielded localizations for each vertebra. 

\begin{figure*}[!t]
    \centerline{\includegraphics[width=0.95\textwidth]{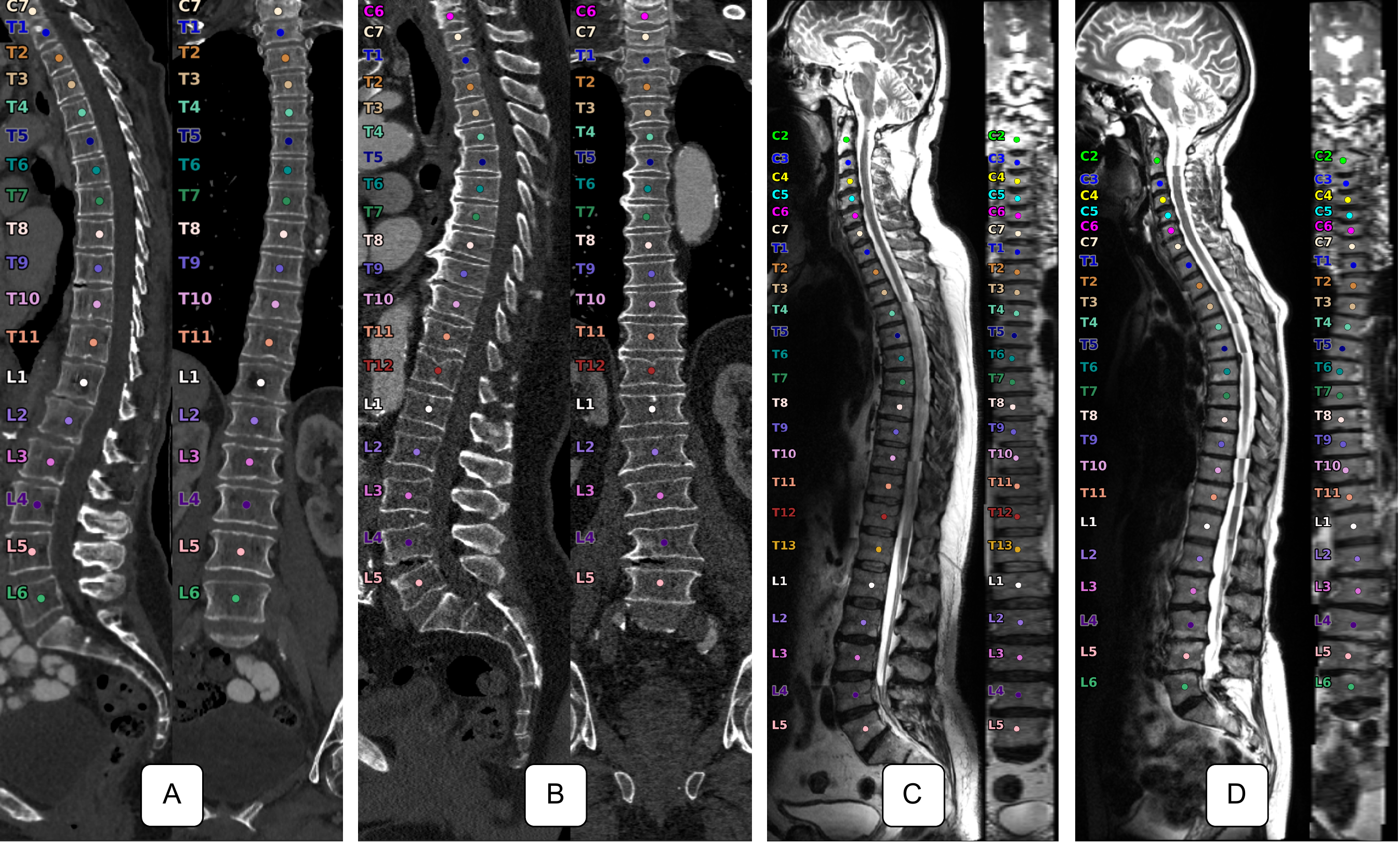}}
    \caption{Example data samples from our CT (A, B) and MRI (C, D) dataset. For each sample, it shows a sagittal and coronal projection along the spine as well as the corpus centroids with vertebra labels associated with each. While the CT images contain various field-of-views, the NAKO T2w sagittal MRI images always cover the whole spine. A) spine with eleven thoracic vertebrae, B) normal spine with twelve thoracic and five lumbar vertebrae, C) has an additional thoracic vertebra, and D), a subject with only eleven thoracic vertebrae but six lumbar ones.}
    \label{fig:datasample}
\end{figure*}

The T2w sagittal images of the NAKO always show the whole spine split into three segments and thus contain up to 26 fully visible vertebrae (see Figure \ref{fig:datasample}). Using the open-source toolbox TPTBox\footnote{https://github.com/Hendrik-code/TPTBox}, we stitched the different segments into one image\cite{graf2025generating}. If we now label the vertebrae in cranial-to-caudal order, we would already be correct except for TEA and LEA cases.

For CT, we utilized an in-house cohort consisting of images from various scanners. Some CT images are contrast-enhanced. The data include segmentation annotations for the vertebrae, supervised by an expert with 24 years of experience. In contrast to the T2w MRI of the NAKO, the FOV of the CTs varied heavily, containing a range of 8 -- 25 fully visible vertebrae (average 16.7 $\pm$ 3.6 vertebrae). We utilized 1536 images in which the first thoracic vertebra is present (see Figure \ref{fig:datasample}). This is necessary to manually assess TEA and create the reference labels.

In both MRI and CT, we consciously split in a way so that the test set contains more TEA and LEA to better evaluate the models.

When manually rating the thoracolumbar junction, we adapted the categorization of Park et al. \cite{park2016thoracolumbar}. We defined a thoracic vertebra as one having a true rib, regardless of the rib's length. If there is only a unilateral or bilateral accessory ossification centre (comparable to Type 4 as defined in \cite{park2016thoracolumbar}), we determined this to be a lumbar vertebra. In ambiguous cases, we judged by the morphology (e.g. facet orientation) whether a vertebra is thoracic.

For the lumbosacral transition, we adapted the analysis of Konin et al. \cite{konin2010lumbosacral}. In the presence of a lumbosacral transitional Castellvi anomaly, we defined the vertebra with this anomaly as a sacralized lumbar vertebra regardless of the Castellvi type. The segmentation model SPINEPS we used on the NAKO data was already trained to make that distinction, which we only had to manually correct in 16 cases ($0.25\%$). Similarly, the annotations of the CT data already distinguished between the lumbar and sacral vertebrae.

One expert with 3 years of experience performed the manual annotations for all data.

\begin{figure*}[!t]
    \centerline{\includegraphics[width=0.95\textwidth]{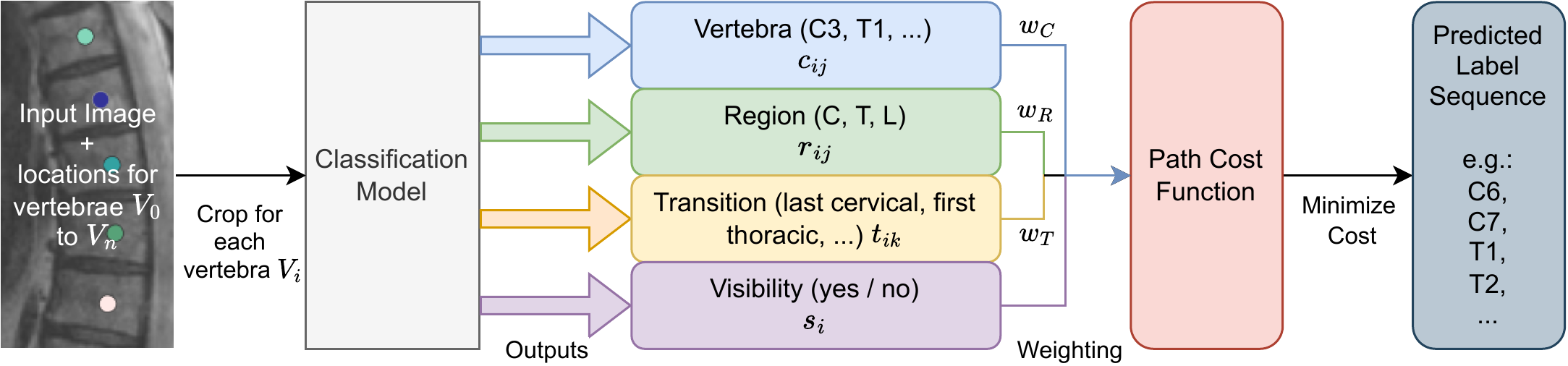}}
    \caption{Overview of the VERIDAH approach. We take an input image and the vertebra localizations to crop around each vertebra. We then input these 3D cropped images separately into our classification model. The classification model has four output heads. We weigh their predictions based on our calibration and combine them into a cost matrix. Finally, an algorithm finds the optimal label sequence path that minimizes this cost, making our final sequence prediction.}
    \label{fig:pipeline}
\end{figure*}

\subsection{VERIDAH}
Our proposed approach consists of two main parts, a classification model and a sequence predictor. The sequence predictor is a constraint optimization algorithm that guarantees a logical sequence of vertebra labels. The input variables for this algorithm are derived per individual vertebra, which are obtained using a classification model. 

\subsubsection{Data Preprocessing}
For training and inference, our approach requires the location of each vertebra in a given image. Since we already have instance segmentation annotations for our data, we utilize those to compute the center of mass locations for each vertebra. With those locations, we calculate crops for each vertebra, where the center of mass location becomes the center. We keep neighboring vertebrae visible to give the classification model context.

We reorient these cutouts consistently and align the resolution depending on whether it is T2w MRI or CT (see Table \ref{tab:hparams}). 

\subsubsection{Classification Model}
We trained a 3D DenseNet169 \cite{huang2017densely} on individual vertebra cutouts. With a typical one-classification head approach, one would train it to output the vertebra label (i.e., C3, T5, \dots). In contrast, VERIDAH utilizes four different classification heads. (1) Target vertebra label (such as C3, T5, \dots) with 23 classes; (2) target region with the three classes cervical, thoracic, and lumbar; (3) and target vertebra transition with six classes (is the vertebra the last cervical, first thoracic, last thoracic, first lumbar, last lumbar, or none above); (4) target visibility (if the vertebra is fully visible or not) as one class output.
We replace the original classification head with the four prediction heads. Each has an additional fully connected linear layer and a ReLU activation function, followed by the output linear layer and the softmax function.

Since especially T13 cases are rare, training directly on class labels like T13 would create a huge class imbalance problem for the classifier. To circumvent this, we replace the T13 and L6 labels in the reference annotation with T12 and L5, respectively. Thus, the existence of a T13 and L6 is indicated by two consecutive T12 or L5 labels. This means that during training, the label difference between a T12 and a T13 is only present to the model through the vertebra transition output head. The T13 is labeled as the last thoracic vertebra; the T12 in a case with 13 thoracic vertebrae is not.

All the vertebrae in our datasets are sufficiently visible to train on. Thus, we introduce the visibility label during training as data augmentation (denoted RandCut). We randomly crop a part of the image, and if the vertebra segmentation is not fully visible anymore, we set the visibility reference label to zero, one otherwise. 
    
We used Pytorch v2.04.1 \cite{imambi2021pytorch} with Pytorch Lightning v2.0.8 \cite{sawarkar2022deep} and the MONAI libraries v1.3.0 \cite{cardoso2022monai}. 

To account for the rare occurrence of these TEA and LEA anomalies, we sample each anomalous case in the training set four times during each epoch instead of only once. For the hyperparameters employed, see Table \ref{tab:hparams}.
    \begin{table}[htbp]
        \centering
        \caption{Training Hyperparameters. Different setups for the sequences are written with slashes (T2w sag / CT)}
        \label{tab:hparams}
        \setlength{\tabcolsep}{3pt}
        \begin{tabular}{rc}
            \toprule
            Parameter & Value \\
            \midrule
            \multicolumn{2}{c}{Data} \\ \gccmidrule{1}{2}
            Number of 3D scans & 5291 / 1171\\
            Orientation & (Posterior, Inferior, Right) \\ 
            Resolution (mm) & (0.85, 0.85, 3.3) / (1.0, 1.0, 1.0) \\ 
            Crop Size (voxel count) & (200, 160, 32) / (128, 128, 128) \\ 
            \midrule
            \multicolumn{2}{c}{Training Setup} \\ 
            \gccmidrule{1}{2}
            Architecture & DenseNet 169 \\
            Batch Size & 32 \\
            Max Epochs & 100 \\
            LR & 1e-4 \\ 
            LR End Factor & 1e-2 \\
            L2 Regularization & 1e-3 \\
            Dropout & 0.1 \\
            \midrule
            \multicolumn{2}{c}{Data Augmentation} \\ 
            \gccmidrule{1}{2}
            RandCut & prob=0.20 \\
            RandAffine & prob=0.40, translate=8, scale=0.15, degrees=15° \\
            RandScaleIntensity & prob=0.15, offset=0.15 \\
            RandShiftIntensity & prob=0.15, factor=0.15 \\
            RandGaussianSmooth & prob=0.10, sigma=0.60 \\
            RandGaussianNoise & prob=0.10, std=10.0 \\
            RandAdjustContrast & prob=0.10, gamma=1.0 \\
            RandLowResolution & prob=0.10, range=(0.8, 1.0) \\
            \bottomrule
        \end{tabular}
    \end{table}

    After classifying a sequence of vertebrae, the output is further processed. We add a Gaussian filter on each softmax output per vertebra and then normalize the vector by dividing it by its length. For the vertebra transition output, we normalize over the different classes (i.e., the more vertebrae the model predicts to be the last thoracic one, the lower the values for each of those predictions).

    Let $V = \{V_1, V_2, \ldots, V_n\}$ be the sequence of $n$ vertebra instances in an image, spatially ordered from top to bottom. $C = \{\text{C1}, \text{C2}, \dots, \text{C7},\text{T1} \dots \text{T12}, \text{L1}, \dots \text{L5}\}$ defines all 24 vertebra labels (without T13 and L6) with $C_i$ corresponding to the $i$-th entry in $C$. We define the set of possible vertebra transitions label $T = \{\text{None}, \text{Last Cervical}, \text{First Thoracic}, \text{Last Thoracic}, \newline\text{First Lumbar}, \text{Last Lumbar}\}$.

    With $i$ corresponding to the $i$-th vertebra instance $V_i$, $j$ as the index for the vertebra label $C_j$, and $k$ as the index for the vertebra transition set $T_k$, then the classifier outputs normalized vertebra class softmax scores $c_{ij}$, normalized region prediction scores $r_{ij}$ (where each region output is mapped onto each vertebra class of that region, e.g. the cervical region output is applied to C1 through C7), normalized vertebra transition scores $t_{ik}$, and a normalized visibility softmax score $s_{i} \in [0,1]$. 

    \subsubsection{Sequence Predictor}
    To get the final vertebra labeling sequence for any field of view, we incorporate global constraints similar to Meng et al. \cite{meng2022vertebrae}. For that, we combine the outputs of our classification model into a cost function. The intuition is to find the sequence with the maximum agreement between the different classification outputs across the vertebrae. For instance, one classification head might indicate a T12, while the region output predicted the vertebra belonging to the lumbar region.

    With $w_C$, $w_R$, $w_T$ being weights for the corresponding classifier output types $c_{ij}$, $r_{ij}$, and $t_{ik}$, we calculate the cost for each instance $V_i$ and class $C_j$ with
    \begin{equation}
        \text{labelcost}(i,j) = (c_{ij} \cdot w_C) + (r_{ij} \cdot w_R) 
    \end{equation}

    This way, we can define a labelcost matrix $L^{n\times24}$ where $L_{ij} = \text{labelcost}(i,j)$.
        
    The vertebra transition cost is utilized differently since it depends on the rest of the path. Given a candidate path $p$ as a sequence of vertebra class index labels $[j_0, j_1, \dots, j_n]$ with $p_i \in C$ being the $i$-th entry of $p$, we define the vertebra transition cost for an individual vertebra $V_i$ as:
    \begin{equation}
        \text{transcondition}(p, i, k) =
            \begin{cases}
                t_{ik} & \text{if } T_{k} \text{ fullfilled in } p \\
                0, & \text{otherwise} 
            \end{cases}
    \end{equation}
    \begin{equation}
        \text{transcost}(p, i) = w_T \cdot \sum_{k\in T} \text{transcondition}(p, i, k)
        \label{eq:transcost}
    \end{equation}

    The classifier predicts the visibility $s_i$, which we utilize to weigh each vertebra. The less visible the model predicts the vertebra to be, the less it impacts the sequence predictor. Additionally, by applying a Gaussian filter, we automatically weigh the boundary vertebrae less. We do this since the classifier has less surrounding information when looking at a boundary vertebra or when the vertebra itself is not fully visible, and we observed less accurate individual predictions there. 
    Thus, we can compute the cost of a whole path by using the visibility output and accumulating the individual costs with
    \begin{equation}
        \text{pathcost}(p) = -
            \sum_{i}^n 
                s_i \cdot \left(
                \text{transcost}(p, i) + L_{i,p_i}
                \right)
        \label{eq:pathcost}
    \end{equation}

    Thus, for all $p \in P$ with $P$ being the set of all possible paths, the sequence predictor solves the minimization problem and finds the solution path $X$ by

    \begin{equation}
        X = \argmin_{p \in P} \text{pathcost}(p)
    \end{equation}

    In order to do so, we utilize dynamic programming and recursively run over all possible path options and compute the cost. 
     However, we can enforce constraints by setting them at an infinite cost. Our constraints, regardless of FOV, are as follows:
    \begin{itemize}
        \item The labels must be consecutive (e.g, T5 must be followed by T6). To account for T11 cases, we allow an L1 directly after a T11.
        \item The labels must be in spatial order (not T6 and then T5).
        \item There can only be a maximum of seven cervical vertebrae, 11 to 13 thoracic vertebrae, and only four to six lumbar vertebrae.
        \item We allow a T12 followed by a T12 exactly once to account for T13, and do the same with two consecutive L5s for L6 cases.
    \end{itemize}
    In contrast to Meng et al. \cite{meng2022vertebrae}, we do not add artificial costs to the paths containing a TEA or LEA.

    Once we have computed $X$, all we need to do is replace the class indices $p_i$ with the actual class labels $C_{p_i}$, then relabel occurrences of two consecutive T12s to [T12, T13] and two consecutive L5s to [L5, L6] to receive the final result. 

    \begin{figure}[htbp]
        \centerline{\includegraphics[width=0.95\linewidth]{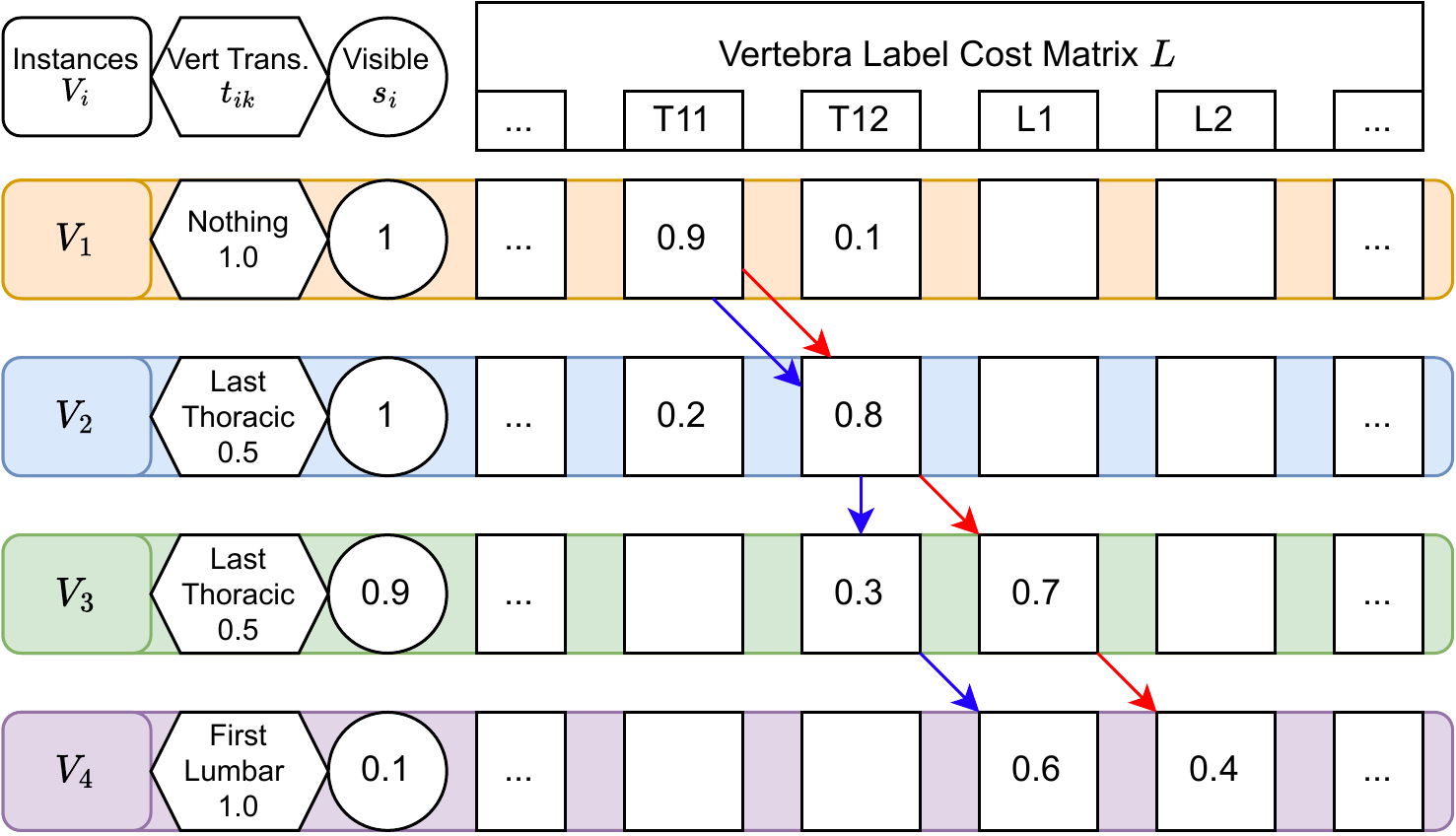}}
        \caption{Example of our sequence predictor on an artificially created sample containing $n=4$ vertebrae ($V_1$ -- $V_4$) in the FOV. The classifier outputs specific vertebra class predictions $c_{ij}$ and region prediction scores $r_{ij}$ for each class (already combined into the label cost matrix $L$ and reduced to T11 to L2 for simplicity), their visibility $s_i$, and the vertebra transitions output $t_{ik}$. Depending on the utilized classification head, either the path [T11, T12, L1, L2] (red arrows) or [T11, T12, T12, L1] (blue arrows) is returned by the sequence predictor. In the case of the latter one, the post-processing relabels the path [T11, T12, T12, L1] to [T11, T12, T13, L1], making this a TEA case. All unspecified classifier outputs have a value of zero.
        }
        \label{fig:path1}
    \end{figure}%

    To demonstrate our approach, consider the example as shown in \ref{fig:path1}. For simplicity, we have already calculated the label cost matrix $L$. Using only the label cost (i.e. putting the transcost $t_{ik}$ to zero and the visibility $s_i$ always to one), the red path of [T11, T12, L1, L2] would be the final sequence, as the red path cost is $-(0.9 + 0.8 + 0.7 + 0.4) = -2.8$ compared to the cost of $-(0.9 + 0.8 + 0.3 + 0.6) = -2.6$ for the blue path. If we consider the vertebra transitions classifier output (Vert Trans.) as well, we need to add $1.0$ to any path where $V_4$ is the L1, and $0.5$ to any path where $V_2$ or $V_3$ is the last thoracic one (see Equation \ref{eq:transcost}). The cost of the red and blue path becomes 
    $-3.3$ and 
    $-4.1$, respectively. The blue path is the final predicted label sequence. If we additionally consider the output of the visibility classification head, $V_4$ has a very low visibility of $s_4 = 0.1$, meaning the vertebra is barely visible, and the remaining classifications are probably uncertain. Thus, all weights coming from $V_4$ are multiplied by $0.1$, and each $V_3$ output is multiplied by $0.9$ (see Equation \ref{eq:pathcost}). If we compare the cost of the blue and red path again using all three outputs, the red path dominates again with a cost of 
    $-2.87$ compared to blue with a cost of 
    $-2.58$.

\section{Experiments}

For both CT and MRI, we individually train a classification model. 
We train the T2w sagittal MRI model on the native resolution of the NAKO images, which is approximately $(0.85,0.85,3.3)$ mm with the orientation being Posterior, Inferior, Right consistently. We resampled the CT images consistently to an isotropic $(1.0, 1.0, 1.0)$ mm resolution.

Since the models solve the same task, we calibrated the sequence predictor weights $w_C$, $w_R$, and $w_T$ based on both models on a $\sim 10\%$ split from the training set. Based on this calibration, we set $w_C = 0.9$, $w_R = 1.1$, and $w_T = 0.6$. We observed these to be quite stable regardless of the split utilized. Then we evaluated the models on the corresponding test sets. 

\subsection{Experimental Setup}

For benchmarking, we evaluate VERIDAH against the approach from Meng et al. \cite{meng2022vertebrae} in CT, while in T2w sagittal MRI, we use SpineNetV2 \cite{windsor2022spinenetv2} and Warszawer et al. \cite{yehudawarszawer202514920281}. Notably, these baselines detect the vertebra locations in the image in addition to labeling. To mitigate propagation errors originating from merged or missed detections and thus perform a fairer comparison to VERIDAH, we replaced the detection process in the Meng et al. algorithm with the ground-truth locations. We removed any wrong locations that were predicted by any of those methods. Thus, we only compare the vertebra labeling algorithms of those methods. Meng et al. introduce an artificial cost to predicting TEA and LEA, which may cause under-prediction of TEA and LEA; thus, we additionally compare against Meng et al. where these penalty costs are set to zero. We denote this Meng et al. ZP.

We train a baseline model for ablation studies using only the vertebra label classification output, representing the naive approach. We then evaluate several configurations on each image sequence independently: (1) a model trained solely on vertebra-level output and predicting labels directly without a sequence predictor (VH-V), (2) a model trained on vertebra-level output but using the sequence predictor constrained to this single output (VH-P), and (3) the complete VERIDAH approach, trained with four classification outputs and utilizing the complete sequence predictor (4H-P). Additionally, to test a naive integration strategy, we construct a combined cost matrix from all classification outputs and assign labels by taking the per-vertebra argmax, bypassing the sequence predictor. We denote this variant as 4H-AM.

As additional ablations, we analyze the performance of VERIDAH if we introduce an artificial cost to the paths containing enumeration anomalies similar to Meng et al. \cite{meng2022vertebrae} (see Section \ref{app:extracost}). 

The proposed VERIDAH method always predicts a sequence that consists of consecutive vertebrae. In practice, faulty methods to extract the positions of the vertebrae or imaging issues (like artifacts or signal loss) might lead to a missing vertebra in the detection process. Thus, we analyze how VERIDAH would perform in these circumstances. First, we ran the same setup on the test data without adding gaps in the reference data, but removing the consecutive constraint on the sequence predictor. This means that the reference data does not contain gaps, but our method can now predict gaps (see Section \ref{app:gaps}). Then we randomly introduce a gap in each test subject by removing one random vertebra from the reference labels. We rerun the same experiment, showing how well VERIDAH performs in correctly predicting gaps.

In practice, the availability of these enumeration anomaly labels is sparse. To showcase the ability of VERIDAH to predict TEA and LEA even though it has never seen those labeled cases during training, we retrain our approach and remove all TEA and LEA annotations by always assuming twelve thoracic vertebrae. This effectively means we relabel all T13 vertebrae to L1 and the L1 of T11 cases to T12 and shift the remaining lumbar vertebra labels accordingly. We evaluate on our unmodified test data, thus including the TEA and LEA labels.

The final ablation is about FOV consistency. We analyzed this by cropping the test images to all permutations of different numbers of visible vertebrae and then comparing the predictions over these cut images against the cut reference labels.

\subsection{Evaluation}

For evaluation, we compare the labeling sequences of predictions with the references. We count the individual labels that were predicted correctly and the number of subjects where the whole sequence was predicted correctly. In detail, we define the \textit{perfect label percentage} (\PLP) as the percentage of subjects for which the labeling was correct for all vertebrae in that subject. We denote the \textit{subject correctness} (\subjcorr) as the average number of correct vertebrae in a subject. To focus on the performance of enumeration anomalies, we also define the TEA recall (\TEAacc) and LEA recall (\LEAacc) as the percentage of correctly identified TEA and LEA, respectively. 
Specifically, we define a correct T11 case if the model predicts the sequence [T11, L1] and the reference has the same labels at the same positions. A correctly labeled T13 case is one where the model predicts two consecutive T12s at matching positions. Correct L4 and L6 cases are identified by both sequences ending on L4 or two consecutive L5, respectively. Notably, other labeling errors apart from the enumeration anomalies are not reflected in these two scores, as they already influence the \PLP and the subject correctness metrics.

We employ the Wilcoxon signed-rank test for statistical significance, with a p-value $< 0.05$ indicating significance.

%% file: LaTeX/m_results.tex
\section{Results}
\label{sec:result}

Training until convergence took roughly one GPU day on an NVIDIA A40.

\subsection{CT}

\begin{figure}[htbp]
    \centerline{\includegraphics[width=0.95\linewidth]{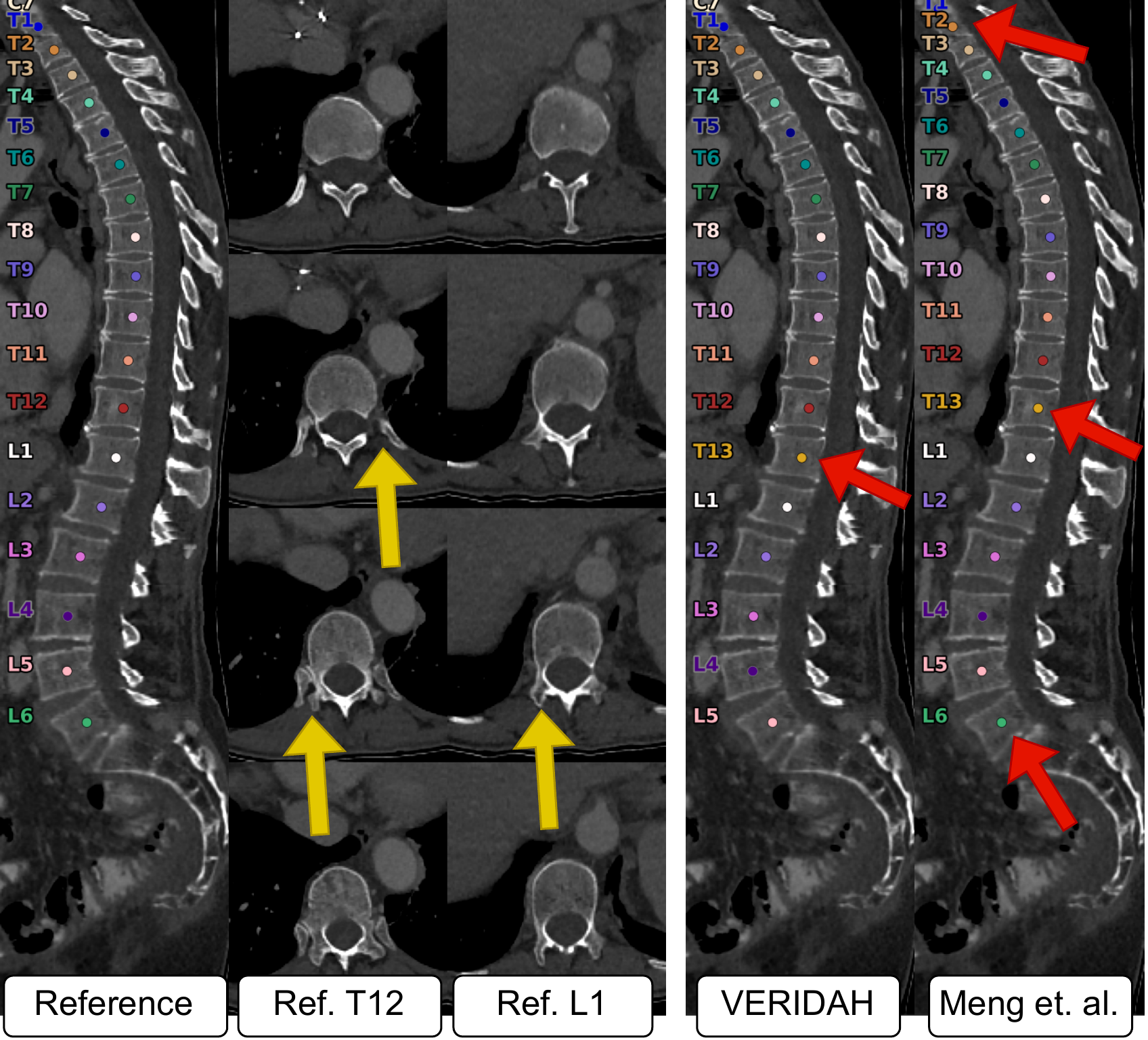}}
    \caption{Example of a CT test subject where all models failed. The reference data (left) shows an L6 LEA case with an ordinary number of twelve thoracic vertebrae. VERIDAH (right) mistakes the L1 for a thoracic vertebra, resulting in thirteen thoracic and five lumbar vertebrae (red arrows). The prediction of Meng et al. has a shift error by one, but correctly labels the L1. The center shows the reference T12 and L1 in multiple axial slices (top to bottom). While the T12 has clear ribs on both sides (yellow arrows), the L1 has thoracic-looking facets while having no ribs, making this a TLTV.}
    \label{fig:ct_pred1}
\end{figure}

\begin{figure}[htbp]
    \centerline{\includegraphics[width=0.95\linewidth]{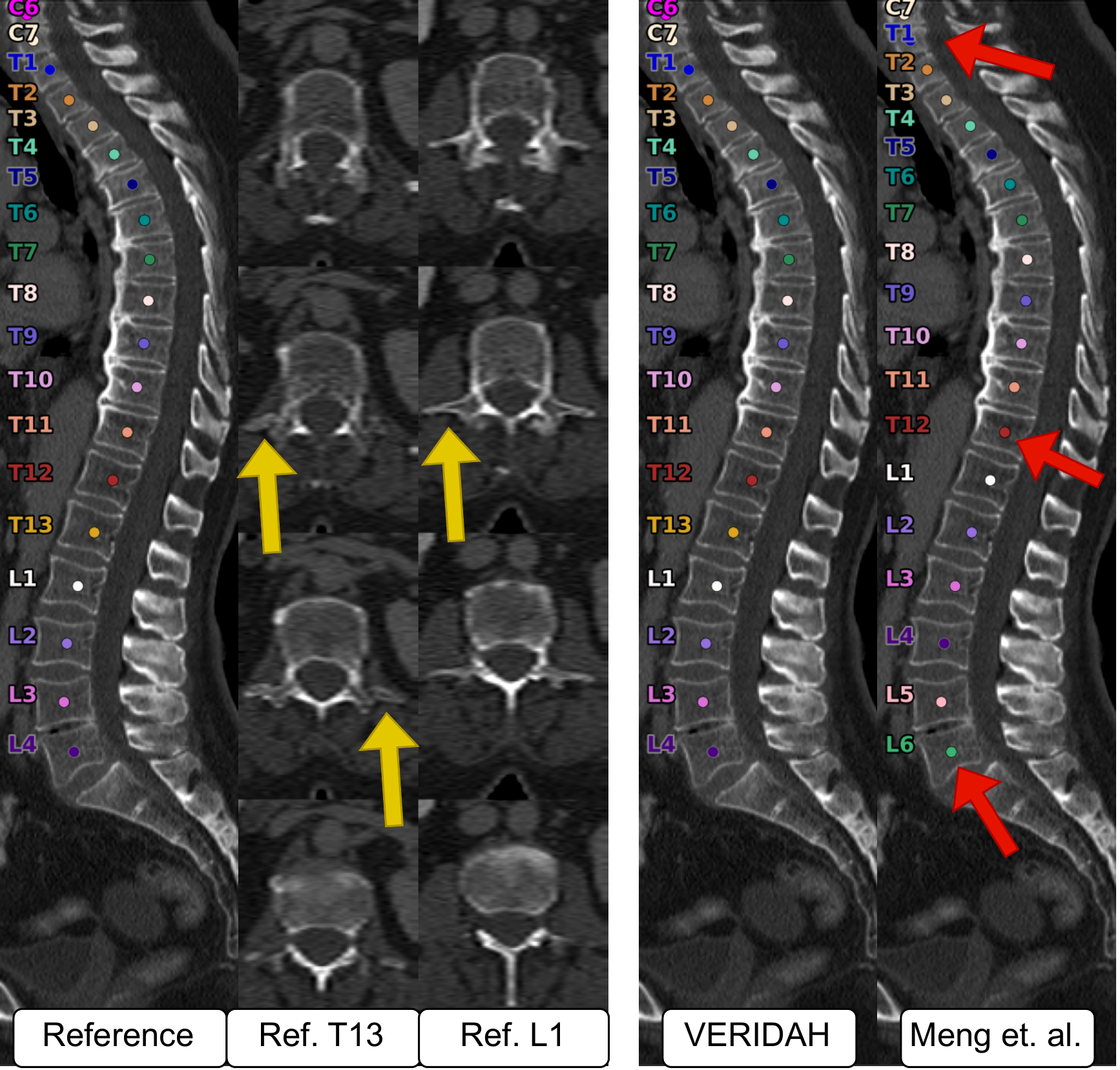}}
    \caption{Example of a CT test subject with a T13 and only four lumbar vertebrae. The axial slices of the reference data (left) show that the T13 has a rib on either side (yellow arrows). While VERIDAH (right) correctly classifies the whole sequence of labels, the prediction of Meng et al. made two major errors, leading to six lumbar vertebrae instead of four (red arrows). }
    \label{fig:ct_pred2}
\end{figure}

VERIDAH achieves significantly better metric values than Meng et al. (see Table \ref{tab:ctresult}, all $p<0.001$). While Meng et al. provide decent results for LEA recall, they fail to correctly annotate TEA cases, regardless of whether we set their anomaly penalty cost to zero. Notably, VERIDAH achieves better performance on non-anomalous cases, as indicated by the superior subject correctness with much higher mean and lower standard deviation. When we set the punishment cost for TEA to zero, their perfect label percentage drops slightly while the TEA recall significantly increases. However, with $14.81\%$ compared to VERIDAH's $96.30\%$, this is still inferior. For a qualitative example, see Figure \ref{fig:ct_pred1}. Notably, on our test set, in contrast to Meng et al., the predicted labels of VERIDAH were never shifted by more than one (see Figure \ref{fig:ct_pred2} for example).

\begin{table}
    \centering
    \caption{Labeling Performance for CT data. Abbreviations: rec: recall, TEA: Thoracolumbar Enumeration Anomaly, LEA: Lumbar Enumeration Anomaly. Meng et al. ZP is the approach from Meng et al., but setting the TEA and LEA penalty costs to zero.}
    \label{tab:ctresult}
    \setlength{\tabcolsep}{3pt}
    \begin{tabular}{rccc}
        \toprule
        Metric              & VERIDAH               & Meng et al. \cite{meng2022vertebrae}        & Meng et al. \cite{meng2022vertebrae} ZP \\
        \midrule
        \PLP             & $\textbf{\VctPLA}$               & $77.81$                    & $77.26$   \\
        \subjcorr          & $\textbf{\VctC}$              & $89.04 \pm 27.73$         &$88.48 \pm 28.28$  \\
        \TEAacc            & $\textbf{\VctTEA}$               & $11.11$            & $14.81$  \\
        \LEAacc            & $\textbf{\VctLEA}$               & $61.11$            &$61.11$ \\
        \bottomrule
    \end{tabular}
\end{table}

\subsection{T2w MRI}

As shown in Table \ref{tab:nakoresult}, VERIDAH outperforms the other baselines in every metric on the NAKO test set (all $p<0.001$). Notably, while both baselines have decent LEA accuracies around $83\%$, they fail to accurately detect TEAs. With VERIDAH, we achieved a perfect labeling sequence in $98.3\%$ of subjects of the test set, while correctly predicting $87.8\%$ and $94.48\%$ of the TEA and LEA cases, respectively. The subject correctness is consistently high across the board with relatively low standard deviation, indicating that erroneous labels mostly involve one or two vertebrae within one subject.

\begin{table}
    \centering
    \caption{Labeling Performance for T2w sagittal MRI data. Abbreviations: rec: recall, TEA: Thoracolumbar Enumeration Anomaly, LEA: Lumbar Enumeration Anomaly.}
    \label{tab:nakoresult}
    \setlength{\tabcolsep}{3pt}
    \begin{tabular}{rccc}
        \toprule
        Metric              & VERIDAH               & SpineNetv2 \cite{windsor2022spinenetv2}        & TotalSpineSeg \cite{yehudawarszawer202514920281} \\
        \midrule
        \PLP             & $\textbf{\VmriPLA}$               & $91.49$           & $94.24$     \\
        \subjcorr          & $\textbf{\VmriC} $  & $97.89 \pm 9.32$  & $98.98 \pm 4.13$    \\
        \TEAacc            & $\textbf{\VmriTEA}$               & $0.0$             & $27.63$       \\
        \LEAacc            & $\textbf{\VmriLEA}$               & $83.72$           & $83.04$       \\
        \bottomrule
    \end{tabular}
\end{table}

%% file: LaTeX/m_ablations.tex
\subsection{Ablations}\label{sec:ablations}

In this section, we conduct ablation studies to analyze our proposed VERIDAH approach in more detail.

\subsubsection{Classification Heads and Sequence Predictor}

Both the multi-classification head approach and our proposed sequence predictor provide significant performance boosts (see Table \ref{tab:nakoabl}, all $p<0.001$). Notably, without the sequence predictor, the added benefit of four classification outputs compared to one (4H-AM vs. VH-V) is barely noticeable. The four classification outputs become more meaningful when combined with the sequence predictor (4H-P vs. VH-P). These behaviors are very similar both in CT and MRI.

\begin{table}
    \centering
    \caption{Ablation Labeling Performance. Abbreviations: rec: recall, TEA: Thoracolumbar Enumeration Anomaly, LEA: Lumbar Enumeration Anomaly, VH-V: Taking only the vertebra labeling without sequence predictor, VH-P: Inputting only the vertebra output to the sequence predictor, 4H-AM: Taking the Argmax of the four outputs instead of the sequence predictor. 4H-P: The proposed approach of utilizing the four classification outputs with the sequence predictor.}
    \label{tab:nakoabl}
    \setlength{\tabcolsep}{3pt}
    \begin{tabular}{rccccc}
        \toprule
        Metric & VH-V & VH-P & 4H-AM & 4H-P\\
        \midrule
        \multicolumn{5}{c}{T2w sagittal} \\ \gcmidrule{5}
        \PLP             & $82.00$ & $95.6$ & $82.10$ & $\textbf{\VmriPLA}$ \\
        \subjcorr          & $97 \pm 6.75$ & $98.92 \pm 5.05$ & $ 97.45 \pm 6.69 $ & $\textbf{\VmriC}$  \\
         \TEAacc            & $56.10$ & $80.49$ & $57.32$  & $\textbf{\VmriTEA}$ \\
        \LEAacc            & $65.52$ & $80.69$ & $67.59$  & $\textbf{\VmriLEA}$\\
        \midrule
        \multicolumn{5}{c}{CT} \\ \gcmidrule{5}
        \PLP             & $73.42$ & $95.62$ & $73.42$ & $\textbf{\VctPLA}$ \\
        \subjcorr          & $95.95 \pm 8.66$ & $98.75 \pm 5.89$ & $95.92 \pm 8.74$ & $\textbf{\VctC}$  \\
        \TEAacc            & $74.07$ & $62.96$ & $74.07$  & $\textbf{\VctTEA}$ \\
        \LEAacc            & $30.56$ & $63.89$ & $30.56$  & $\textbf{\VctLEA}$\\
        \bottomrule
    \end{tabular}
\end{table}

\subsubsection{Extra Cost Weights}\label{app:extracost}
\begin{figure*}[!htbp]%
    \centering
    \begin{subfigure}{0.5\textwidth}%
    \includegraphics[width=0.98\linewidth]{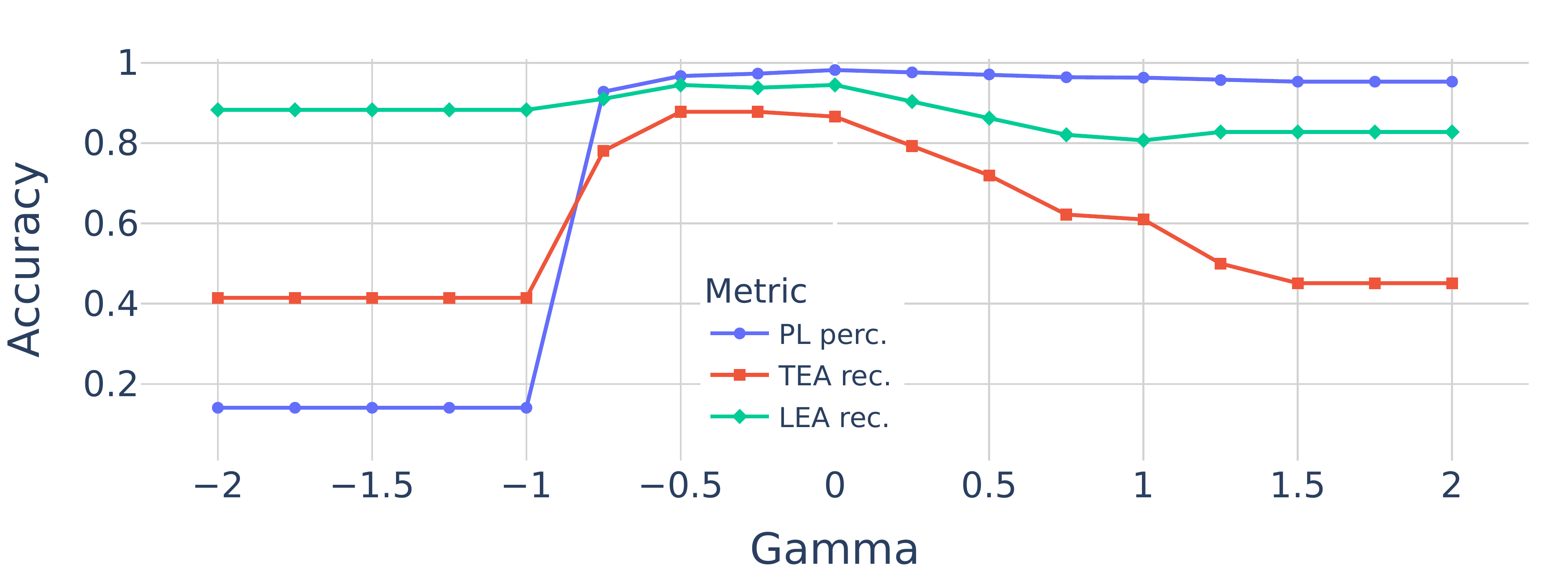}
    \caption{MRI}
    \end{subfigure}%
    \begin{subfigure}{0.5\textwidth}%
    \includegraphics[width=0.98\linewidth]{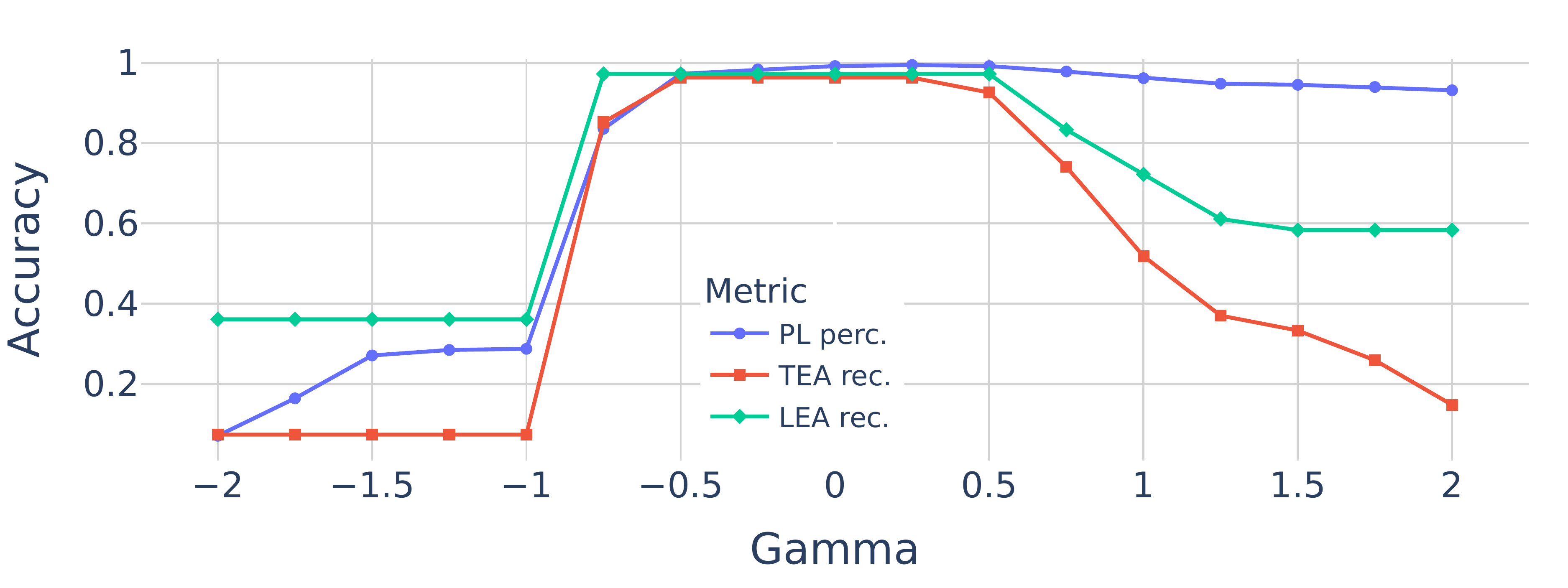}
    \caption{CT}
    \end{subfigure}%
    \caption{The behavior of three metrics in relation to a specified artificial cost punishment of $\gamma$ on our test data. We sample these results in $0.25$ steps, as indicated by the plot markers. The baseline is represented by $\gamma = 0$.}
    \label{fig:ablteacost}
\end{figure*}

Meng et al. \cite{meng2022vertebrae} add an artificial penalty to the prediction of TEA and LEA cases, making it less likely to predict these. We investigate this by introducing a cost change of $\gamma$ for paths containing a TEA or LEA during our sequence predictor. A positive $\gamma$ makes the model predict more TEA and LEA, while a negative cost introduces more difficulty for the model to predict anomalous sequences. With $\gamma \in [-2, 2]$, we report the changes in metrics using VERIDAH on our test data in Figure \ref{fig:ablteacost}.

At $\gamma \le -1.0$, there is a huge drop in performance. This is probably because at this point, the model only predicts anomalous sequences. Since most subjects have no enumeration anomalies, the sequence predictor has catastrophic performance. With $\gamma \ge 1.25$, the cost of anomalies is so high that it rarely predicts anomalous sequences, thus leading to lower TEA and LEA recall.

Introducing any such cost (regardless of positive or negative) negatively influences the sequence predictor. There is one notable exception in MRI. With $\gamma = -0.25$, there is a slight increase in the TEA recall. from $86.59$ to $87.8$ while the \PLP decreases from $98.2$ to $97.4$. We deem this tradeoff unworthy. We thus refrain from using such a cost in our main approach.

\subsubsection{Vertebra Gaps}\label{app:gaps}

\begin{figure*}[!htbp]%
    \centering
    \begin{subfigure}{0.5\textwidth}%
    \includegraphics[width=0.98\linewidth]{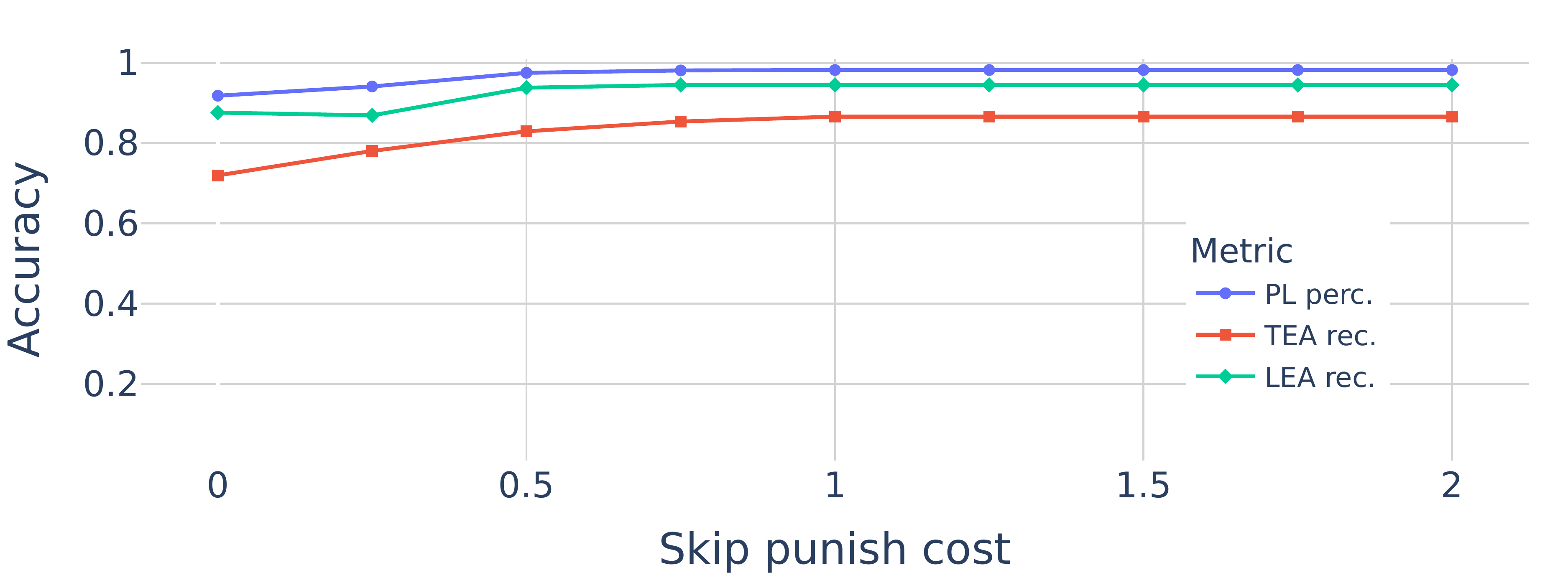}
    \caption{MRI}
    \end{subfigure}%
    \begin{subfigure}{0.5\textwidth}%
    \includegraphics[width=0.98\linewidth]{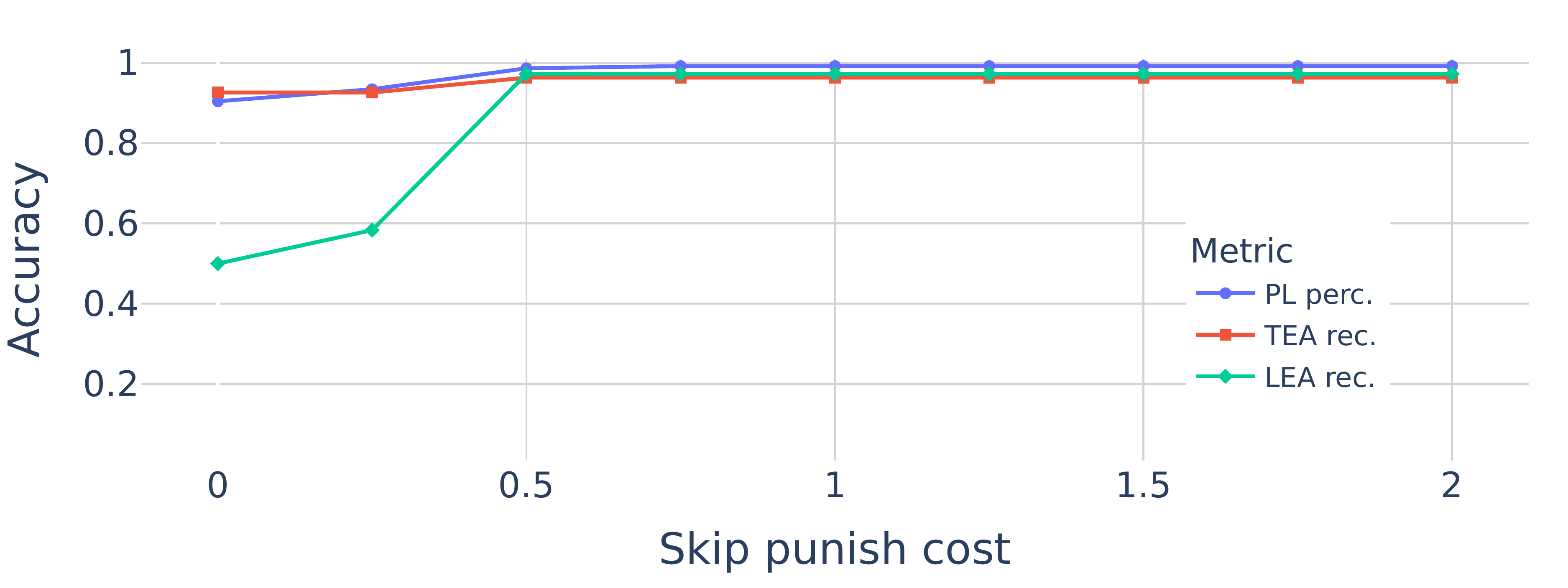}
    \caption{CT}
    \end{subfigure}%
    \caption{The behavior of three metrics, if we allow arbitrary label gaps in our sequence predictor on the test data. The x-axis shows different costs to punish these skips. We sample these results in $0.25$ steps, as indicated by the plot markers.}
    \label{fig:ablskipcost}
\end{figure*}

Figure \ref{fig:ablskipcost} highlights how an artificial penalty for each gap in predicted sequences compares to the performance on our reference data. 

With a punishment cost of $\ge 1.0$, the model never predicts any gaps on the MRI test data. Thus, performance is unchanged. The lower the cost, the worse the performance across all three metrics. This indicates that allowing our approach to predict non-consecutive label sequences leads to predicting gaps where none exist.

\begin{figure*}[!htbp]%
    \centering
    \begin{subfigure}{0.5\textwidth}%
    \includegraphics[width=0.98\linewidth]{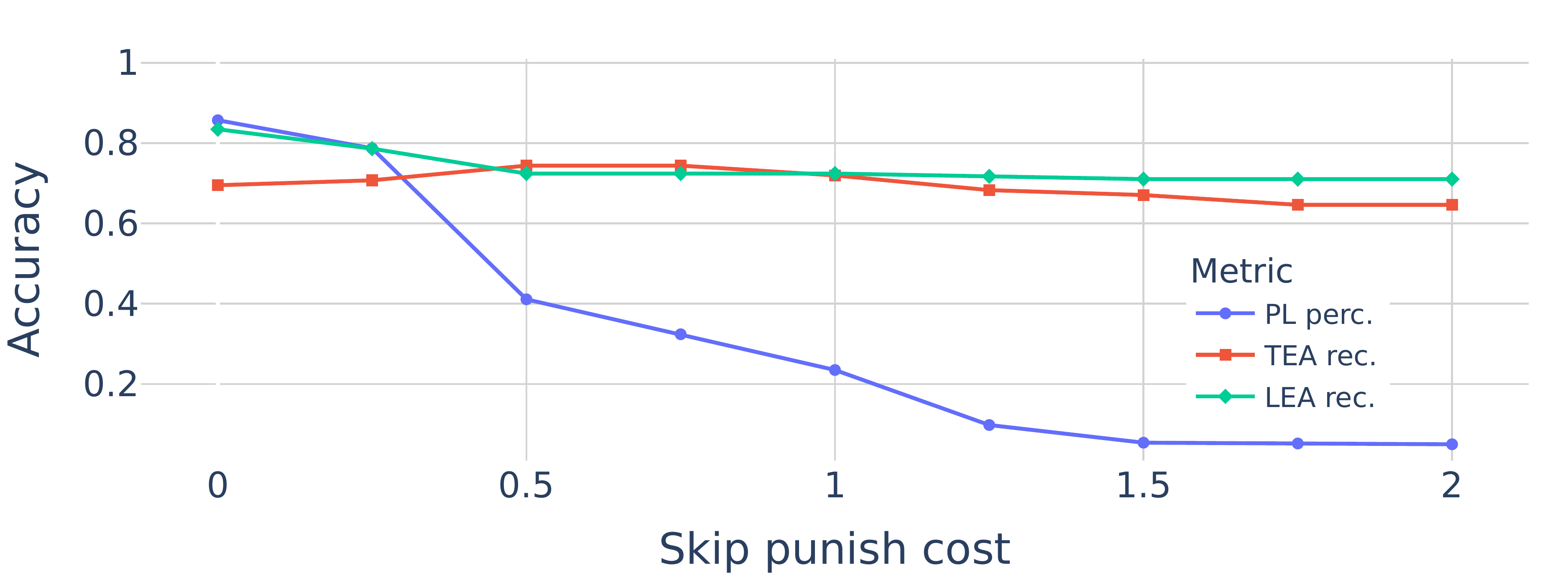}
    \caption{MRI}
    \end{subfigure}%
    \begin{subfigure}{0.5\textwidth}%
    \includegraphics[width=0.98\linewidth]{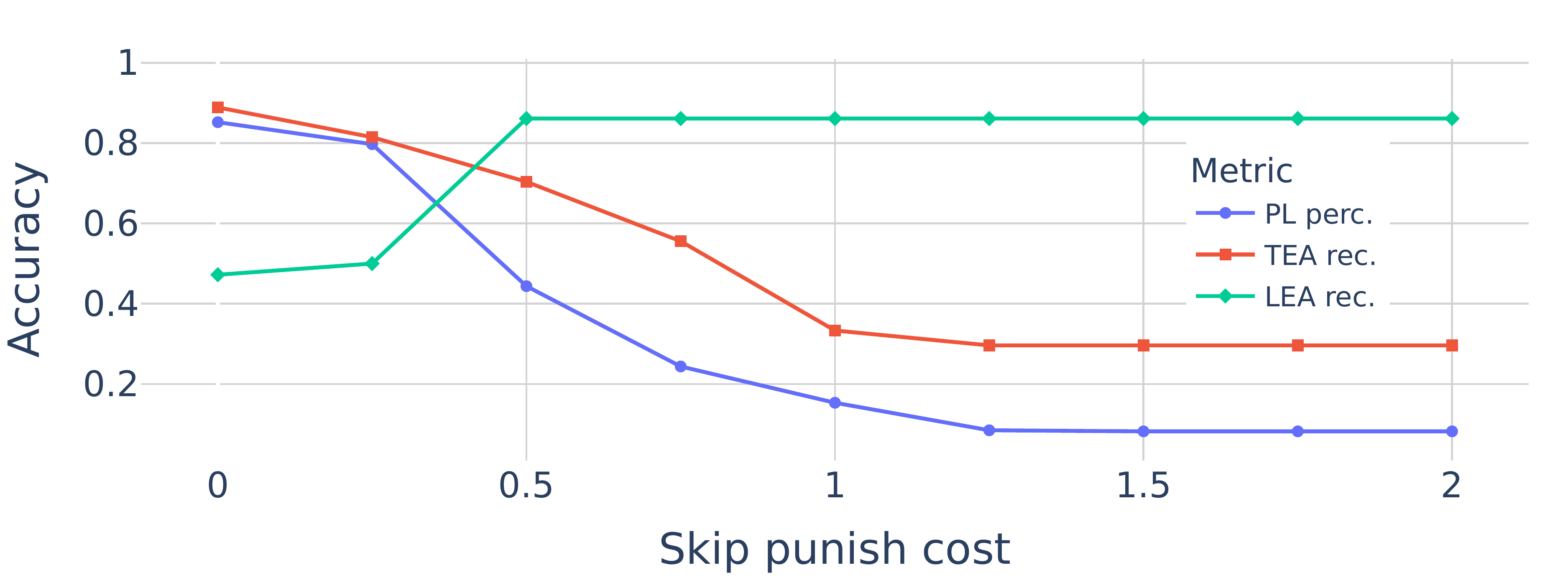}
    \caption{CT}
    \end{subfigure}%
    \caption{The behavior of three metrics, if we allow arbitrary label gaps in our sequence predictor on the test data, and we remove exactly one random vertebra in each subject. The x-axis shows different costs to punish these skips. We sample these results in $0.25$ steps, as indicated by the plot markers.}
    \label{fig:ablrealskipcost}
\end{figure*}

When we introduce random but fixed gaps into our MRI test data by just removing vertebra locations from the reference and re-running this experiment, we achieve a different result (see Figure \ref{fig:ablrealskipcost}). We observe that VERIDAH achieves optimal results if we keep the skip penalty to zero in the case of actual gaps in the data. The performance drops significantly once we punish the model for predicting gaps in the label sequence, since each subject contains gaps. Interestingly, the \TEAacc has a small peak around the interval $[0.75, 1.00]$ in the MRI sequence, for which we can offer no explanation other than random noise. 

All in all, this means that even if subjects have missing vertebrae, if we allow VERIDAH to predict gaps in the sequence, it is correct (including the gaps) in $85.7\%$ of MRI subjects, with a \TEAacc of $69.51\%$ and \LEAacc of $83.45\%$. For CT, VERIDAH achieves perfect labels in $85.21\%$ of subjects, with a \TEAacc of $88.89\%$ and \LEAacc of $47.22\%$.

\subsubsection{Train without TEA and LEA annotations}\label{app:noanomalies}

The metrics we achieved with VERIDAH trained on zero correctly annotated TEA and LEA cases are reported in Table \ref{tab:nakoablna}. We again compare using all four classification heads with the sequence predictor (4H-P-NA) against using only the vertebra classification output with the sequence predictor (VH-P-NA).

\begin{table}
    \centering
    \caption{Labeling Performance T2w sagittal compared to training without labeled anomalies. Abbreviations: rec: recall, TEA: Thoracolumbar Enumeration Anomaly, LEA: Lumbar Enumeration Anomaly. 4H-P-NA: Taking all four classification heads with the sequence predictor using no TEA or LEA labels during training. VH-P-NA: Using only the vertebra output and the sequence predictor using no TEA or LEA labels during training.}
    \label{tab:nakoablna}
    \setlength{\tabcolsep}{3pt}
    \begin{tabular}{rccc}
        \toprule
        Metric              & VERIDAH               & 4H-P-NA        & VH-P-NA \\
        \midrule
        \multicolumn{4}{c}{T2w sagittal} \\ \gcmidrule{4}
        \PLP             & $\VmriPLA$               & $93.40$           & $93.10$     \\
        \subjcorr          & \VmriC            & $ 98.32 \pm 6.52 $      & $ 98.22 \pm 6.96 $    \\
        \TEAacc            & $\VmriTEA$               & $32.93$             & $39.02$       \\
        \LEAacc            & $\VmriLEA$               & $82.76$           & $77.24$       \\
        \midrule
        \multicolumn{4}{c}{CT} \\ \gcmidrule{4}
        \PLP             & $\VctPLA$               & $93.70$           & $92.60$     \\
        \subjcorr          & \VctC             & $ 98.08 \pm 7.77 $  & $ 97.78 \pm 08.24 $    \\
        \TEAacc            & $\VctTEA$               & $25.93$             & $22.22$       \\
        \LEAacc            & $\VctLEA$               & $63.89$           & $52.78$       \\
        \bottomrule
    \end{tabular}
\end{table}

This shows that even if we relabel our reference to count to twelve thoracic vertebrae, VERIDAH still detects about a third of the TEA. This highlights how our method is able to derive the important aspects from the different classification outputs despite not having seen any correctly labeled anomalous cases during training.

\subsubsection{Arbitrary FOV Consistency}\label{app:fov}

\begin{figure*}[!htbp]%
    \centering
    \begin{subfigure}{0.5\textwidth}%
    \includegraphics[width=0.98\linewidth]{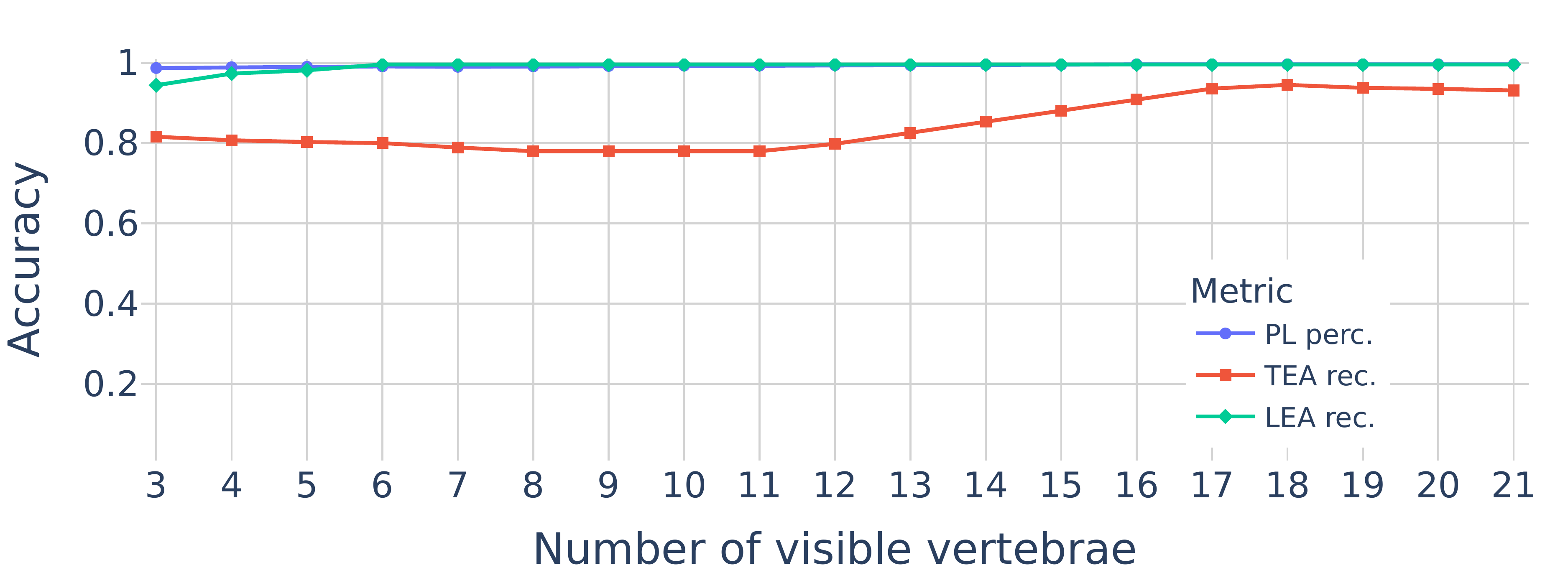}
    \caption{MRI}
    \end{subfigure}%
    \begin{subfigure}{0.5\textwidth}%
    \includegraphics[width=0.98\linewidth]{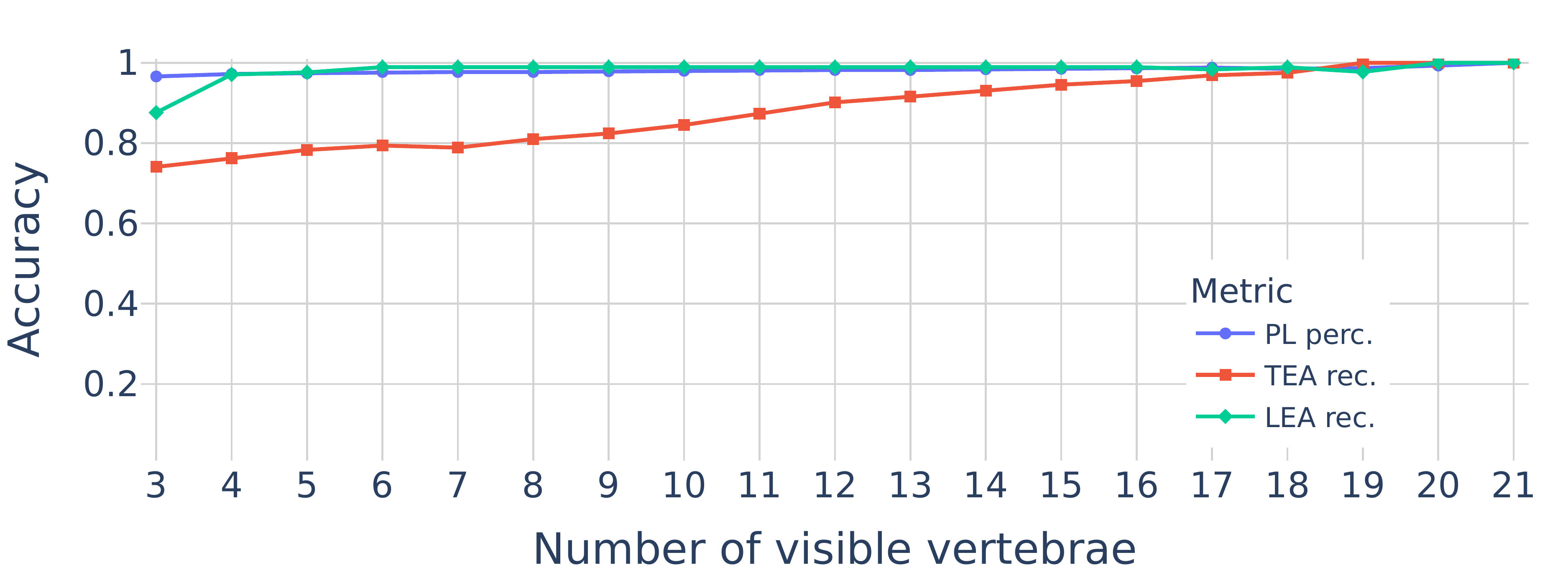}
    \caption{CT}
    \end{subfigure}%
    \caption{The behavior of three metrics over different FOVs of the test data. The x-axis shows the number of visible vertebrae fed into VERIDAH, while the y-axis showcases the different metric values.}
    \label{fig:ablfov}
\end{figure*}%

Figure \ref{fig:ablfov} highlights the robustness of VERIDAH on arbitrary FOVs. The perf. label perc. metric is consistently high, while only the TEA recall suffers from small FOVs. We also observed that mostly a couple of samples were misclassified as soon as an individual vertebra was present in the FOV. This leads to the reason that individual vertebrae are challenging for VERIDAH, not specific FOVs. 

When we relate how consistent VERIDAH's prediction is in any FOV compared to VERIDAH's prediction on the whole spine, there are virtually no inconsistencies except for a handful of permutations of samples out of the test samples. We observed that these are mostly the same samples and FOVs that are falsely predicted from VERIDAH in those limited FOVs.

%% file: LaTeX/m_discussion.tex
\section{Discussion and Conclusion}
\label{sec:discussion}


We demonstrate VERIDAH, a 3D vertebra labeling approach that combines multiple classification heads with a constraint optimization solver to predict vertebra labels. Our approach performs better than state-of-the-art T2w MR and CT imaging methods. VERIDAH works on arbitrary FOVs and can consistently label enumeration anomalies. 

Prior to this study, no approach had been able to reliably label these anomalies, which were often overlooked in studies. We showed that a model trained without an adequate number of correctly labeled enumeration anomalies performs significantly worse. Based on this, we hypothesize that a vertebra labeling methodology can never perform above a certain threshold without considering enumeration anomalies, since the TEA and LEA cases inherently contradict the non-anomalous labels. Significant future improvements in vertebra labeling will likely be achieved mainly through access to more data and advanced, fine-grained definitions of TEA and LEA.

Automatically identifying the correct vertebra labels in any CT or MRI image is of high clinical relevance to provide consistent, robust assessments while reducing radiologists' workload. In many smaller FOV images (especially the ones without a visible T1 or sacrum), we observed that manual annotation is more error-prone and inconsistent than an automated system. 

Nevertheless, our study has its limitations. The most important one to address is the requirement for vertebral localizations. VERIDAH does not perform a localization or detection step prior to labeling, but instead requires the centroids of vertebrae beforehand. However, the method SPINEPS \cite{moller2024spineps}, which we used for our MRI data, had a localization error in $0.53\%$ of subjects. Public datasets such as VerSe \cite{sekuboyina2021verse} exist with similar automated detection approaches for CT. Therefore, as publicly available models can perform localization through segmentation, this requirement should not hinder fellow researchers and clinicians. Nevertheless, we do acknowledge the increased effort required to effectively utilize VERIDAH.

Since the VERIDAH model trained on T2w was only trained on the NAKO cohort, we cannot guarantee the same performance on T2w TSE MRI from a different scanner. Since the CT data used is very diverse, VERIDAH should perform more robustly in that modality.

Overall, we believe that VERIDAH provides a foundation that bridges the gap to enumeration-anomaly-aware labeling. Since many studies analyze different aspects of the spine in relation to vertebra labels, and our approach is publicly available and offers more accurate labels, VERIDAH opens up more detailed research opportunities for the future. 